\documentclass[letterpaper, 10 pt, conference]{ieeeconf}  % Comment this line out if you need a4paper

\IEEEoverridecommandlockouts                              % This command is only needed if 
\usepackage{graphics} % for pdf, bitmapped graphics files
\usepackage{epsfig} % for postscript graphics files
\usepackage{times} % assumes new font selection scheme installed
\usepackage{amsmath} % assumes amsmath package installed
\usepackage{amssymb}  % assumes amsmath package installed

\usepackage{algorithm}
\usepackage{algpseudocode}
\usepackage{subcaption}
\usepackage{cleveref}
\usepackage{cite}
\usepackage{placeins}
\usepackage{afterpage}
\usepackage{xcolor}

\title{\LARGE \bf
PC-SRIF: Preconditioned Cholesky-based Square Root Information Filter for Vision-aided Inertial Navigation
}

\author{Tong Ke$^{\dag}$, Parth Agrawal$^{\dag}$, Yun Zhang$^{\dag}$, Weikun Zhen$^{\dag}$, Chao X. Guo$^{\dag}$, Toby Sharp$^{\dag}$ and Ryan C. Dutoit$^{\dag}$% <-this % stops a space
\thanks{$^{\dag}$Authors are with Google XR \texttt{\{tongk, agrawalparth, zhayun, weikunz, chaoguo, 
tobysharp, rdutoit\}@google.com}.}%
}

\renewcommand{\baselinestretch}{0.99}

\begin{document}

\maketitle
\thispagestyle{empty}
\pagestyle{empty}
%\pagestyle{plain} % TODO: USE empty for submission!!!!!!!!!!!!!!!!!!!!!!!!!!!!!!!!!!!!!!!!!!!!!!!!!!!!!!!!!!!!!!!!!!!!!!!!

%%%%%%%%%%%%%%%%%%%%%%%%%%%%%%%%%%%%%%%%%%%%%%%%%%%%%%%%%%%%%%%%%%%%%%%%%%%%%%%%
\begin{abstract}
In this paper, we introduce a novel estimator for vision-aided inertial navigation systems (VINS), the Preconditioned Cholesky-based Square Root Information Filter (PC-SRIF). When solving linear systems, employing Cholesky decomposition offers superior efficiency but can compromise numerical stability. Due to this, existing VINS utilizing (Square Root) Information Filters often opt for QR decomposition on platforms where single precision is preferred, avoiding the numerical challenges associated with Cholesky decomposition. While these issues are often attributed to the ill-conditioned information matrix in VINS, our analysis reveals that this is not an inherent property of VINS but rather a consequence of specific parameterizations. We identify several factors that contribute to an ill-conditioned information matrix and propose a preconditioning technique to mitigate these conditioning issues. Building on this analysis, we present PC-SRIF, which exhibits remarkable stability in \textit{performing Cholesky decomposition in single precision when solving linear systems in VINS}. Consequently, PC-SRIF achieves superior theoretical efficiency compared to alternative estimators. To validate the efficiency advantages and numerical stability of PC-SRIF based VINS, we have conducted well controlled experiments, which provide empirical evidence in support of our theoretical findings. Remarkably, in our VINS implementation, PC-SRIF's runtime is 41\% faster than QR-based SRIF.
\end{abstract}
\section{Introduction}
A vision-aided inertial navigation system (VINS), which combines data from inertial measurement units (IMU) and cameras, has been a primary solution for 6 degrees-of-freedom (DOF) pose (position and orientation) estimation tasks~\cite{Huang2019a}.
Given its low cost, independence from external infrastructure such as GPS, and high accuracy, VINS has led to numerous successful applications, including augmented reality (AR), virtual reality (VR), and robotics. In these applications, it is essential to provide accurate pose estimation with minimal latency, typically on a platform with limited computing resources, such as VR headsets and AR glasses.
While existing VINS estimators have demonstrated their capability on these platforms, VINS remains an intensive perception task.
Thus, it is beneficial to further enhance the efficiency of VINS estimators, so as to conserve battery power and provide low-latency pose estimates with minimal computational resource contention. In what follows, we first provide a concise overview of existing VINS estimators and then focus on those which prioritize efficiency.

The estimation problem in VINS is a sequential one, where the goal is to produce a trajectory, a sequence of 6 DOF poses, given IMU and camera image input sequences. Disregarding the sequential structure and attempting to solve for all poses simultaneously leads to a nonlinear batch least-squares (BLS) problem under reasonable assumptions.
BLS methods (e.g.,~\cite{Dellaert2006,Triggs1999,Thrun2006,Konolige2008,demmel2021squareba}) achieve high accuracy, but their computational cost increases at least linearly with time due to the growing size of input and output.
Therefore, they are primarily utilized in offline processing rather than real-time applications. To reduce time complexity to constant level in sequential estimation, filtering methods are commonly employed. Particularly, most filters for VINS recursively estimate the probability distribution of a sliding window of recent states, which fall under the category of Bayes filters. VINS based on filtering have demonstrated real-time performance with acceptable accuracy.
Additionally, recent systems incorporate a multi-threaded framework (e.g.,~\cite{Qin2018,campos2021orb,Liu2018}), where a BLS estimator (or an approximated one) operates concurrently with a filter, to integrate loop closure information and mitigate long-term drift.
Nonetheless, the efficiency of the filter remains critical in such systems for producing poses with minimal latency.

In Bayes filters, although it is possible to impose few assumptions regarding the probability distribution of the states (e.g., particle filter), popular VINS filters assume the Gaussian distribution to attain high efficiency, leading to systems based on the extended Kalman filter (KF\footnote{Henceforth, we omit “extended” and employ the same acronym to refer to both the extended and standard versions of these filters.}) (e.g.,~\cite{Mourikis2007,geneva2020openvins,Li2013,van2023eqvio,Bloesch2017,fan2024schurvins}) or its variants, such as the extended information filter (IF)~\cite{huang2011observability}.
KF exhibits sufficient numerical stability to enable single precision (float32) for VINS~\cite{peng2024ultrafast}. However, as~\cite{BarShalom2004} has noted, the computational cost can be reduced by replacing KF with IF, when the dimension of the measurements $m$ is large compared to the dimension of the states $n$, a common scenario in VINS.

IF employs the information matrix rather than the covariance matrix to parameterize the Gaussian distribution, and most optimization-based VINS (e.g., \cite{Qin2018,Leutenegger2015,von2018direct,chen2023optimization}) can be categorized as utilizing IF, in the sense that their prior information from marginalization includes an information matrix.
% ORB-SLAM3 (campos2021orb) should not be here since they do not always require a prior.
% [better stability can be achieved by another alternative, the square root KF, while often at a higher computational cost; at best, \cite{peng2024ultrafast} achieves $O(3mn^2)$ in the update step] 
%
For VINS, IF can attain lower complexity if the problem structure is appropriately exploited.
Specifically, when $m$ significantly surpasses $n$, IF offers a notable computational advantage over other variants, attributed to employing the Cholesky solver for the linear least-squares (LS) problem arising in the update step [$O(1mn^2)$ floating point operations (FLOPs)].
However, IF has been criticized for its numerical instability. In particular, \cite{Wu2015} asserted that the information matrix in VINS is often ill-conditioned, necessitating double precision (float64) for IF. Thus, the square root information filter (SRIF) is employed instead~\cite{Wu2015, demmel2021square,givens2023square, huai2022square, huai2024consistent}, where the QR decomposition is used for solving the LS, requiring more FLOPs [$O(2mn^2)$] but sufficiently stable to operate in float32 for VINS.

Among all aforementioned methods, the most efficient way in FLOPs to implement the update step of VINS is using the Cholesky LS solver in IF, but the necessity for float64 makes it less efficient than the QR solver in float32 SRIF in practice~\cite{demmel2021square}.
In this paper, however, we reveal that the information matrix in VINS is not inherently ill-conditioned but contingent upon the parameterization of the states.
Based on this finding, we propose a preconditioning technique that effectively reparameterizes the states so that the Cholesky LS solver is well-conditioned in float32.
Consequently, the update step is theoretically more efficient than that of float32 SRIF. Furthermore, to take another advantage of SRIF, only a sub-block of the square root information matrix requires updating at each step~\cite{Kaess2008}, we suggest utilizing the Cholesky-based update in SRIF instead of IF. These advancements lead to a novel SRIF-based estimator for VINS, which we designate as the Preconditioned Cholesky-based Square Root Information Filter (PC-SRIF). As evident later, PC-SRIF achieves greater efficiency than existing filters for VINS without compromising accuracy. Our main contributions are:
\begin{itemize}
    \item To the best of our knowledge, we present the first work that explains and addresses the conditioning issues associated with the information matrix in VINS. This enables the fast Cholesky LS solver in float32.
    \item We propose a novel estimator, termed PC-SRIF, which effectively leverages our analysis of the conditioning issue and leads to a highly efficient VINS.
    \item We introduce a faster algorithm for SRIF marginalization, further enhancing the efficiency of (PC-)SRIF.
    \item We conduct a well controlled experiment on VINS using KF, SRIF, and PC-SRIF to demonstrate that PC-SRIF achieves the highest efficiency among these filters.
\end{itemize}

\section{Information Filters Review and PC-SRIF}\label{sec:estimator}
In this section, we review IF/SRIF and formally describe the algorithm of PC-SRIF as a general estimator. Specifically, we concentrate on the update step, as (i) it is the most computationally expensive operation when the dimension of measurements $m$ significantly exceeds that of states $n$, and (ii) PC-SRIF and SRIF differ solely in this step.

To begin, recall that IF and SRIF recursively estimate the distribution of states represented by $\mathbf{x}\sim\mathcal{N}(\hat{\mathbf{x}}, \mathcal{I}^{-1})$ and $\mathbf{x}\sim\mathcal{N}(\hat{\mathbf{x}}, (\mathbf{R}^\top\mathbf{R})^{-1})$, respectively, where $\mathbf{x}\in\mathbb{R}^n$ and the square root information matrix $\mathbf{R}$ is an upper triangular Cholesky factor of the information matrix $\mathcal{I}$. At each time step, we perform the update given a linearized measurement model $\mathbf{z} | \mathbf{x}\sim\mathcal{N}(\mathbf{r}+\mathbf{H}(\mathbf{x}-\hat{\mathbf{x}}),\mathbf{I})$, where $\mathbf{z},\mathbf{r}\in\mathbb{R}^m,\mathbf{H}\in\mathbb{R}^{m\times n}$ are the measurement, measurement residual, and measurement Jacobian matrix, respectively (without loss of generality, assuming identity measurement noise covariance since we can always whiten the noise). To determine the posterior distribution $\mathbf{x} | \mathbf{z}$, we employ the Maximum a Posteriori estimator and yield a LS system:
\begin{align}
     \min_{\scriptscriptstyle\mathbf{x}}&\left\|\mathbf{R}(\mathbf{x}-\hat{\mathbf{x}})
    \right\|^2 +\left\|\mathbf{H}(\mathbf{x}-\hat{\mathbf{x}})-\mathbf{r}
    \right\|^2\\
    \Leftrightarrow
    \min_{\scriptscriptstyle\delta\mathbf{x}}&\left\|\begin{bmatrix}
    \mathbf{R} \\
    \mathbf{H}
    \end{bmatrix}
    \delta\mathbf{x}-\begin{bmatrix}\mathbf{0}\\\mathbf{r}\end{bmatrix}
    \right\|^2\ (\delta\mathbf{x}\triangleq\mathbf{x}-\hat{\mathbf{x}})\label{eq:update_ls}
\end{align}
In IF, \eqref{eq:update_ls} is solved by constructing the normal equation \eqref{eq:normal_eq}, where a Cholesky factorization of the posterior information matrix $\mathcal{I}^\oplus$ is followed to obtain the solution.
\begin{align}
    \mathcal{I}^\oplus\delta\mathbf{x}=\mathbf{H}^\top\mathbf{r}\ \quad (\mathcal{I}^\oplus=\mathbf{R}^\top\mathbf{R}+\mathbf{H}^\top\mathbf{H}=\mathcal{I}+\mathbf{H}^\top\mathbf{H}) \label{eq:normal_eq}
\end{align}
In contrast, SRIF begins with a QR factorization \eqref{eq:ls_qr}, subsequently solving a squared linear system \eqref{eq:ls_qr2}.
\begin{align}
    \begin{bmatrix}
    \mathbf{R} \\
    \mathbf{H}
    \end{bmatrix}=\mathbf{Q}\begin{bmatrix}\mathbf{R}^\oplus\\\mathbf{0}\end{bmatrix},&\ 
    \begin{bmatrix}\mathbf{r}^\oplus\\\mathbf{e}\end{bmatrix}\triangleq\mathbf{Q}^\top\begin{bmatrix}\mathbf{0}\\\mathbf{r}\end{bmatrix} \label{eq:ls_qr}\\
 \mathbf{R}^\oplus\delta\mathbf{x}&=\mathbf{r}^\oplus \label{eq:ls_qr2}
\end{align}
Notably, when $m$ dominates $n$, the major computational cost of IF is the construction of $\mathcal{I}^\oplus$, with the matrix multiplication requiring $O(2mn^2)$ FLOPs (the Cholesky factorization takes $O(\frac{1}{3}n^3)$ FLOPs and thus  is not major). In SRIF, the most computationally demanding operation is the QR factorization, also requiring $O(2mn^2)$ FLOPs. However, because $\mathcal{I}^\oplus$ is symmetric, only approximately half of its elements need to be computed. Consequently, IF actually requires only $O(1mn^2)$ FLOPs, approximately twice as efficient as SRIF.

Despite the advantages of the IF, it suffers from reduced numerical stability. In the aforementioned steps, the information matrix $\mathcal{I}^\oplus$ is inverted, and the condition number of this matrix is the square of that of $\mathbf{R}^\oplus$ in SRIF. Consequently, for ill-conditioned systems, IF requires the utilization of floating-point numbers with higher precision than SRIF, potentially offsetting their FLOPs difference in practice.

Furthermore, the measurement models in practical systems often involve only a subset of all the states (e.g., the update models of VINS do not include most IMU-related states like IMU biases). In such cases, $\mathbf{H}$ can be expressed in the form of $[\mathbf{0}\ \mathbf{H}_{\scriptscriptstyle2}]$ ($\mathbf{H}_{\scriptscriptstyle2}\in\mathbb{R}^{m\times n_{\scriptscriptstyle2}}, n_{\scriptscriptstyle2}<n$).
In SRIF, if we partition $\mathbf{R},\delta\mathbf{x}$ accordingly and rewrite the cost function as \eqref{eq:srif_small}, we can reduce the problem to another LS \eqref{eq:srif_small1}.
\begin{align}
    &\left\|\begin{bmatrix}
    \mathbf{R}_{\scriptscriptstyle11} & \mathbf{R}_{\scriptscriptstyle12} \\
    \mathbf{0} & \mathbf{R}_{\scriptscriptstyle22}\\
    \mathbf{0} & \mathbf{H}_{\scriptscriptstyle2}
    \end{bmatrix}
    \begin{bmatrix}\delta\mathbf{x}_{\scriptscriptstyle1}\\\delta\mathbf{x}_{\scriptscriptstyle2}\end{bmatrix}-
    \begin{bmatrix}\mathbf{0} \\ \mathbf{0} \\\mathbf{r}\end{bmatrix}
    \right\|^2 \label{eq:srif_small}\\
    \min_{\delta\mathbf{x}_{\scriptscriptstyle2}}&\left\|\begin{bmatrix}
    \mathbf{R}_{\scriptscriptstyle22} \\
    \mathbf{H}_{\scriptscriptstyle2}
    \end{bmatrix}
    \delta\mathbf{x}_{\scriptscriptstyle2}-\begin{bmatrix}\mathbf{0}\\\mathbf{r}\end{bmatrix}
    \right\|^2 \label{eq:srif_small1}
\end{align}
Then, $\delta\mathbf{x}$ can be solved following \eqref{eq:srif_small2} and \eqref{eq:srif_small3}, and the posterior square root information matrix is given by \eqref{eq:srif_small4}. Note that the cost of QR factorization is reduced to $O(2mn_{\scriptscriptstyle2}^2)$. Consequently, in such systems, SRIF offers not only numerical stability but also computational efficiency advantages.
\begin{align}
    \begin{bmatrix}
    \mathbf{R}_{\scriptscriptstyle22} \\
    \mathbf{H}_{\scriptscriptstyle2}
    \end{bmatrix}&=\mathbf{Q}\begin{bmatrix}\mathbf{R}^\oplus_{\scriptscriptstyle22}\\\mathbf{0}\end{bmatrix},\ 
    \begin{bmatrix}\mathbf{r}^\oplus_{\scriptscriptstyle2}\\\mathbf{e}\end{bmatrix}\triangleq\mathbf{Q}^\top\begin{bmatrix}\mathbf{0}\\\mathbf{r}_{\scriptscriptstyle2}\end{bmatrix} \label{eq:srif_small2}\\
 \delta\mathbf{x}_{\scriptscriptstyle2}&=\mathbf{R}_{\scriptscriptstyle22}^{\oplus-1}\mathbf{r}^\oplus_{\scriptscriptstyle2},\ \delta\mathbf{x}_{\scriptscriptstyle1}=-\mathbf{R}_{\scriptscriptstyle11}^{-1}\mathbf{R}_{\scriptscriptstyle12}\delta\mathbf{x}_{\scriptscriptstyle2}\label{eq:srif_small3}\\
 \mathbf{R}^\oplus&=\begin{bmatrix}
    \mathbf{R}_{\scriptscriptstyle11} & \mathbf{R}_{\scriptscriptstyle12} \\
    \mathbf{0} & \mathbf{R}_{\scriptscriptstyle22}^\oplus
    \end{bmatrix}\label{eq:srif_small4}
\end{align}

To combine this advantage of SRIF and IF's efficiency benefit attributed to the Cholesky solver, we introduce a novel estimator, PC-SRIF, which leverages the Cholesky solver within the SRIF framework. To comprehend its operation, note that the normal equation of the LS subproblem~\eqref{eq:srif_small1} is \eqref{eq:normal_eq_small}, which, while solvable directly, can lead to numerical issues similar to those encountered in IF. PC-SRIF addresses this challenge by incorporating preconditioners prior to solving the normal equation. Specifically, \eqref{eq:normal_eq_small} is transformed into \eqref{eq:normal_eq_small_pc} by selecting a preconditioning matrix $\mathbf{M}$ that reduces the condition number of the system to a level where the Cholesky decomposition can be performed with stability.
\begin{align}
    (\mathbf{R}_{\scriptscriptstyle22}^\top\mathbf{R}_{\scriptscriptstyle22}+\mathbf{H}_{\scriptscriptstyle2}^\top\mathbf{H}_{\scriptscriptstyle2})\delta\mathbf{x}_{\scriptscriptstyle2}&=\mathbf{H}_{\scriptscriptstyle2}^\top\mathbf{r}\label{eq:normal_eq_small}\\
     (\mathbf{R}_{\scriptscriptstyle22p}^\top\mathbf{R}_{\scriptscriptstyle22p}+\mathbf{H}_{\scriptscriptstyle2p}^\top\mathbf{H}_{\scriptscriptstyle2p})\mathbf{M}\delta\mathbf{x}_{\scriptscriptstyle2}&=\mathbf{H}_{\scriptscriptstyle2p}^\top\mathbf{r}\label{eq:normal_eq_small_pc}\\
    \mathbf{R}_{\scriptscriptstyle22p}=\mathbf{R}_{\scriptscriptstyle22}\mathbf{M}^{-1},\mathbf{H}_{\scriptscriptstyle2p}&=\mathbf{H}_{\scriptscriptstyle2}\mathbf{M}^{-1}\nonumber
\end{align}
Subsequently, the LS subproblem can be solved using the Cholesky solver following
\begin{align}
    \mathbf{R}_{\scriptscriptstyle22p}^\top\mathbf{R}_{\scriptscriptstyle22p}+\mathbf{H}_{\scriptscriptstyle2p}^\top\mathbf{H}_{\scriptscriptstyle2p}&\stackrel{\text{Cholesky}}{=\joinrel=\joinrel=}\mathbf{R}_{\scriptscriptstyle22p}^{\oplus\top}\mathbf{R}_{\scriptscriptstyle22p}^\oplus\label{eq:normal_eq_small_chol}\\
    \mathbf{R}_{\scriptscriptstyle22}^\oplus=\mathbf{R}_{\scriptscriptstyle22p}^\oplus\mathbf{M}, \quad
    \delta\mathbf{x}_{\scriptscriptstyle2}&=\mathbf{R}_{\scriptscriptstyle22}^{\oplus-1}\mathbf{R}_{\scriptscriptstyle22p}^{\oplus-\top}\mathbf{H}_{\scriptscriptstyle2p}^\top\mathbf{r}\label{eq:dx2_pc}
\end{align}
%in accordance with \eqref{eq:normal_eq_small_chol}-\eqref{eq:dx2_pc}. 
%
By employing this approach, PC-SRIF achieves a significant reduction in the complexity [$O(mn^2_{\scriptscriptstyle2})$ FLOPs], while being as stable as SRIF.
Note that in above steps, \eqref{eq:normal_eq_small_pc} should be directly constructed, instead of forming the worse-conditioned \eqref{eq:normal_eq_small} and then applying preconditioners to it.
This is because the formation of the information matrix in \eqref{eq:normal_eq_small} can result in significant irreversible information loss~\cite[Chapter~5]{golub2012matrix}.
For similar reasons, applying preconditioning to an IF (PC-IF) does not effectively address the conditioning issues, which is another point of using SRIF as our framework.

% Based on the preceding analysis, in this paper, we put forward a novel estimator, PC-SRIF, which effectively combines the computational efficiency advantages of both SRIF and IF. Notably, PC-SRIF utilizes the Cholesky LS solver within the SRIF framework, resulting in a significant reduction in the computational complexity of the update step [$O(mn^2_{\scriptscriptstyle2})$ FLOPs]. However, it is important to acknowledge that this method inherits a numerical issue similar to IF, necessitating the inversion of a well-conditioned matrix w.r.t. the employed machine precision. To address this challenge, PC-SRIF incorporates preconditioners prior to solving the LS problem, effectively reducing the condition number and ensuring the numerical stability of the Cholesky LS solver.

The primary motivation behind introducing preconditioners is to enable the Cholesky LS solver, thereby outperforming the QR LS solver employed by SRIF. Consequently, the computational cost associated with the preconditioners in PC-SRIF must be sufficiently low to not offset the benefits offered by the LS solver.
%E.g., while it is possible to utilize $\mathbf{R}_{\scriptscriptstyle22}$ as the preconditioner, the resulting overhead would outweigh any potential advantages.
Notably, general preconditioners often lack computational efficiency or effectiveness in improving conditioning, compared to those employing domain knowledge.
In the subsequent section, we present our PC-SRIF based VINS, where we leverage VINS specific information to design effective preconditioners and ensure numerical stability in float32.
%%%%%%%%%%%%%%%%%%%%%%%%%%%%%%%%%%%%%%%%%%%%%%%%%%%%%%%%%%%%%%%%%%%%%%%%%%%%%%%%
% \begin{table}[h]
% \caption{An Example of a Table}
% \label{table_example}
% \begin{center}
% \begin{tabular}{|c||c|}
% \hline
% One & Two\\
% \hline
% Three & Four\\
% \hline
% \end{tabular}
% \end{center}
% \end{table}

%   \begin{figure}[thpb]
%       \centering
%       \framebox{\parbox{3in}{We suggest that you use a text box to insert a graphic (which is ideally a 300 dpi TIFF or EPS file, with all fonts embedded) because, in an document, this method is somewhat more stable than directly inserting a picture.
% }}
%       %\includegraphics[scale=1.0]{figurefile}
%       \caption{Inductance of oscillation winding on amorphous
%       magnetic core versus DC bias magnetic field}
%       \label{figurelabel}
%   \end{figure}

% Figure Labels: Use 8 point Times New Roman for Figure labels. Use words rather than symbols or abbreviations when writing Figure axis labels to avoid confusing the reader. As an example, write the quantity ÒMagnetizationÓ, or ÒMagnetization, MÓ, not just ÒMÓ. If including units in the label, present them within parentheses. Do not label axes only with units. In the example, write ÒMagnetization (A/m)Ó or ÒMagnetization {A[m(1)]}Ó, not just ÒA/mÓ. Do not label axes with a ratio of quantities and units. For example, write ÒTemperature (K)Ó, not ÒTemperature/K.Ó

\section{Vision-Aided Inertial Navigation System}
In this section, we present a VINS based on the proposed PC-SRIF. Initially, we provide a brief description of the estimated states and measurement models. Subsequently, we focus on the two key contributions that enhance efficiency against existing SRIF systems: (i) A fast marginalization algorithm that leverages sparsity; (ii) The aforementioned update using the Cholesky LS solver, which is made possible by addressing the conditioning issue in VINS.
\subsection{Estimated States}
Our VINS recursively estimates a state vector, which at time step $k$ is in the form of
\begin{align}
    \mathbf{x}_{state}&=
    \begin{bmatrix}
    {{\mathbf{b}^\top_{g_k}}\ {{\mathbf{b}^\top_{a_k}}}\ {}^{\scriptscriptstyle G}{\mathbf{v}}_{\scriptscriptstyle I_k}^\top\ {{\mathbf{x}}^\top_{slam}} \  t_{sync}\ {{\mathbf{x}}^\top_{poses}}\ {{\mathbf{x}}^\top_{cam}}}
    \end{bmatrix}^\top \\
    \mathbf{x}_{slam}&=
    \begin{bmatrix}
    %{^{\scriptscriptstyle C_{r_1}}\mathbf{f}_1^\top} & {^{\scriptscriptstyle C_{r_2}}\mathbf{f}_2^\top}\dots{^{\scriptscriptstyle C_{r_s}}\mathbf{f}_s^\top}
    \mathbf{f}_1^\top\ \mathbf{f}_2^\top\ \dots\ \mathbf{f}_s^\top
    \end{bmatrix}^\top \\
    \mathbf{x}_{poses}&=
    \begin{bmatrix}
    {}^{\scriptscriptstyle G}{\mathbf{p}}_{\scriptscriptstyle I_{k-l+1}}^\top\ 
    {}^{\scriptscriptstyle G}_{\scriptscriptstyle I_{k-l+1}}\mathbf{q}^\top\ 
    \dots\ 
    {}^{\scriptscriptstyle G}{\mathbf{p}}_{\scriptscriptstyle I_k}^\top\
    {}^{\scriptscriptstyle G}_{\scriptscriptstyle I_k}\mathbf{q}^\top
    \end{bmatrix}^\top\label{eq:state_pose} \\
    \mathbf{x}_{cam}&=
    \begin{bmatrix}
    \zeta^\top\ 
    {}^{\scriptscriptstyle I}\mathbf{p}_{\scriptscriptstyle C}^\top\
    {}^{\scriptscriptstyle I}_{\scriptscriptstyle C}\mathbf{q}^\top
    \end{bmatrix}^\top
\end{align}
where $\{G\}$, $\{I\}$, and $\{C\}$ are the global, IMU, and camera frame respectively, and may be sub-scripted by time steps. $\mathbf{b}_{g_k},\mathbf{b}_{a_k}$ are the biases of gyroscope and accelerometer at time step $k$.
${}^{\scriptscriptstyle G}{\mathbf{v}}_{\scriptscriptstyle I_k}$ is the velocity of the IMU w.r.t $\{G\}$.
$\mathbf{x}_{poses}$ is the sliding window of size $l$ consisting of IMU poses w.r.t $\{G\}$, where $\mathbf{q}$ is the orientation quaternion and $\mathbf{p}$ is the position.
$\mathbf{x}_{slam}$ contains $s$ visual features represented in the inverse depth parameterization~\cite{Civera2008}, where each of them is w.r.t a camera frame in the sliding window.
When a past pose slides out the window, the features w.r.t to that frame are reparameterized to the latest frame. $t_{sync}$ is the time offset between the IMU and camera clock.
$\mathbf{x}_{cam}$ consists of camera-IMU extrinsics and camera intrinsics $\zeta\in\mathbf{R}^4$, focal lengths and camera center. 
% Hence, the corresponding error state is defined as:
% \begin{equation}
% \tilde{\textbf{X}}_I=
% \begin{bmatrix}
% {{{\tilde{\textbf{b}_g}}}^\mathsf{T}\ {{\tilde{\textbf{b}_a}}}^\mathsf{T}\ {^G{\tilde{\textbf{v}}}_I}^\mathsf{T}\ {{\tilde{\textbf{X}}}_{slam}}^\mathsf{T} \  {{\tilde{\textbf{t}}}_{sync}}^\mathsf{T}\ {{\tilde{\textbf{X}}}_{imu}}^\mathsf{T}\ {{\tilde{\textbf{X}}}_{cam}}^\mathsf{T}}
% \end{bmatrix}^\mathsf{T}
% \end{equation}
% \begin{equation}
% \tilde{\textbf{x}}_{imu}=
% \begin{bmatrix}
% {^{G}_{I}\tilde{\boldsymbol{\theta}}}^\mathsf{T}&{^G{\tilde{\textbf{p}}}_I}^\mathsf{T}\cdot\cdot\cdot\cdot \ 
% \end{bmatrix}^\mathsf{T}
% \end{equation}
% \begin{equation}
% \tilde{\textbf{x}}_{camera}=
% \begin{bmatrix}
% {^{G}_{I}\tilde{\boldsymbol{\theta}}}^\mathsf{T}&{^G{\tilde{\textbf{p}}}_I}^\mathsf{T}
% \end{bmatrix}^\mathsf{T}
% \end{equation}
% The corresponding error state in PC-SRIF is defined for each of these elements, except for the quaternion, which utilizes a manifold error definition. All other error states employ the additive error definition.
% \begin{equation}
% \begin{matrix}
%   \textbf{q}^G_I={\delta{^G_I\textbf{q}}} \otimes {^G_I\hat{\textbf{q}}}, & 
%   {\delta{^G_I\textbf{q}}} = 
%   \begin{bmatrix}
%       1 & \frac{1}{2}\delta{^G_I{\tilde{\boldsymbol{\theta}}}}
%   \end{bmatrix}^\mathsf{T}
% \end{matrix}
% \end{equation}

Note that in (PC-)SRIF, the order of the states in $\mathbf{x}_{state}$ has a major impact on efficiency and thus is carefully chosen.
In particular, for the update step, based on what is discussed in Sec.~\ref{sec:estimator}, we can partition the states based on if they are involved in the measurement model so as to minimize $n_{\scriptscriptstyle 2}$.
%, the size of $\delta\mathbf{x}_{\scriptscriptstyle 2}$.
%
For this reason, $\mathbf{b}_{g_k},\mathbf{b}_{a_k},{}^{\scriptscriptstyle G}{\mathbf{v}}_{\scriptscriptstyle I_k}$, which are not used in visual measurements are put on the top of $\mathbf{x}_{state}$.
In addition, as what will be shown in Sec.~\ref{ssec:marginalization}, states with a smaller index is cheaper to marginalize than those closer to the bottom of $\mathbf{x}_{state}$.
Thus, we order the features in ${\mathbf{x}}_{slam}$ and the poses in ${\mathbf{x}}_{poses}$ using chronological order so that states marginalized earlier have smaller indexes.
%The order of different quantities in the state vector plays a significant role on the performance of the filtering steps. This is because propagation and marginalization steps entail matrix operations on the top left corner of the square root information factor R. Conversely, update state involves operations on the bottom right corner of the factor R. With this in mind, the state quantities ( for example, imu biases) that play a major role in propagation step are kept on the leftmost side, and quantities that are more heavily involved in the update step (for example, camera extrinsics) are placed on the rightmost side of the state vector. Moreover, the policy dictates that the new states are always appended to the right side of the vector such that the relative order between different quantities is upheld as outlined in eq [10]. 
\subsection{Measurement Models}
Our VINS utilizes the IMU measurements for propagation and visual measurements for update.
The measurement models and information management employed are similar to those in \cite{Wu2015} and thus we omit their details.
At each time step, our VINS evaluates these measurement functions and their Jacobian matrices, and employ them to perform propagation, marginalization, and update in succession, which are detailed in following subsections.

\subsection{Propagation and State Augmentation}
During the propagation step, new states such as the new pose and new features are added to the state vector.
The square root information matrix $\mathbf{R}$ is modified accordingly similar to \cite{Wu2015}.
However, in contrast to~\cite{Wu2015}, which leaves the augmented $\mathbf{R}$ non-upper-triangular and relies on a Householder QR in the following marginalization step, we perform Givens QR to make $\mathbf{R}$ upper-triangular before marginalization.
As what the next subsection shows, using Givens QR to keep $\mathbf{R}$ upper-triangular is essential for utilizing sparsity and ensuring efficiency.
\subsection{Marginalization}\label{ssec:marginalization}
Marginalization removes states no longer involved in the measurement models (past poses and features losing track) from the state vector, providing a cap for the state size.
In SRIF, a prevalent strategy for modifying $\mathbf{R}$ in marginalization is permuting the column to marginalize in $\mathbf{R}$ to the leftmost, performing a QR decomposition to restore the upper triangular structure, and subsequently removing marginalized state from the top-left of $\mathbf{R}$.
In~\cite{Wu2015,demmel2021square}, the Householder QR is employed, which leads to a time complexity of $O(np^2)$ for marginalizing a scalar state at index $p$.
In this work, we show that it is actually possible to achieve $O(np)$ by using Algo.~\ref{alg:marginalization} (usually multiple states need marginalizing, and we apply Algo.~\ref{alg:marginalization} to them in turn).
The intuition behind this is that we employ the Givens QR to exploit the existing upper triangular structure of $\mathbf{R}$, in contrast to the Householder QR that treats the input as dense.
Remarkably, by employing this approach, our marginalization costs $O(n^2)$ as oppose to $O(n^3)$ with the Householder QR at the worst case ($p\approx n$).
\begin{figure}
    \centering
    \includegraphics[width=0.48\textwidth]{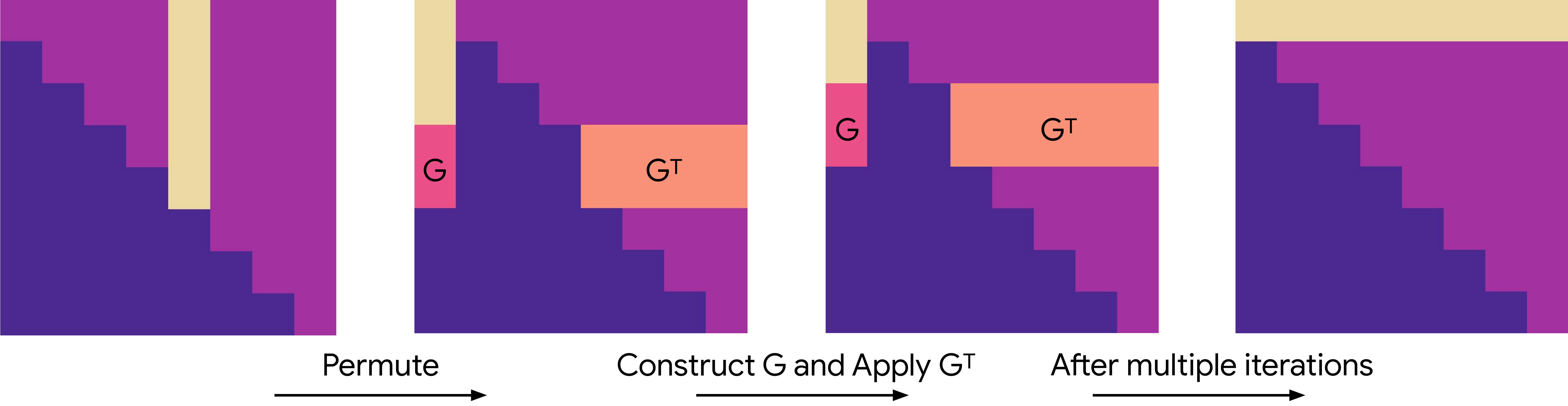}
    \caption{Visualization of our SRIF marginalization algorithm.
    %on $\mathbf{R}$.
    }
    \label{fig:marginalization}
    \vspace{-2mm}
\end{figure}
\begin{algorithm}
\caption{SRIF Marginalization (also see Fig.~\ref{fig:marginalization})}
\begin{algorithmic} 
\State \textbf{Objective:} Marginalize a scalar $x_m$ from a state vector $\mathbf{x}=[\mathbf{x}_{r_1}^\top \ x_m \  \mathbf{x}_{r_2}^\top]^\top$ and modify its square root information matrix $\mathbf{R}$ accordingly while keeping upper-triangular.

\State \textbf{Inputs:} $\mathbf{x}, \mathbf{R}$, $p$ (index of $x_m$ in $\mathbf{x}$).
%, i.e., $\mathbf{x}_{r_1}\in\mathbb{R}^{p-1}$
\State \textbf{Steps:}

\If{$p \neq 1$}
    \State Permute $\mathbf{R}(1:p, p)$ with $\mathbf{R}(1:p,1:p-1)$.
    \For{$j = p$ \textbf{down to} $2$}
        \State Construct Givens rotation $\mathbf{G}$ using $\mathbf{R}(j-1:j,1)$.
        \State Apply $\mathbf{G}^\top$ to $\mathbf{R}(j-1:j,1)$, zeroing out $\mathbf{R}(j,1)$.
        \State Apply $\mathbf{G}^\top$ to $\mathbf{R}(j-1:j, j:end)$.
\EndFor
\EndIf
\State $\mathbf{x} \gets [\mathbf{x}_{r_1}^\top \  \mathbf{x}_{r_2}^\top]^\top$
\State $\mathbf{R} \gets \mathbf{R}(2:end,2:end)$

\end{algorithmic} 
\label{alg:marginalization}
\end{algorithm}
\subsection{Update and Conditioning Analysis}
In the update step, visual measurements are employed to obtain the posterior state estimates.
In Sec.~\ref{sec:estimator}, we have discussed a few alternatives to perform the update in information based filters, and concluded that the one based on the Cholesky decomposition in PC-SRIF yields the fewest FLOPs in theory.
Nonetheless, in VINS literature, the information matrix is widely believed to be too ill-conditioned to perform a Cholesky factorization in float32, which has been a primary motivation for systems employing SRIF and the QR LS solver. In this subsection, however, we state that this is not an inherent property of VINS and present our solution to enable the Cholesky LS solver in float32.

Recall that the condition number $\kappa(\mathbf{A})$ of a matrix $\mathbf{A}$ equals the ratio of its largest to its smallest singular value, $\sigma_{max}(\mathbf{A})/\sigma_{min}(\mathbf{A})$. For the covariance matrix $\mathbf{P}=\mathcal{I}^{-1}$, which has the same condition number as $\mathcal{I}$, these two singular values correspond to the two directions with the greatest and smallest variances, respectively. Consequently, a small $\kappa(\mathbf{P})$ implies that $\mathbf{x}_{state}$ has similar variances in all directions, which is unlikely the case for most VINS state parameterizations.

\begin{figure}
    \centering
    \includegraphics[width=0.48\textwidth]{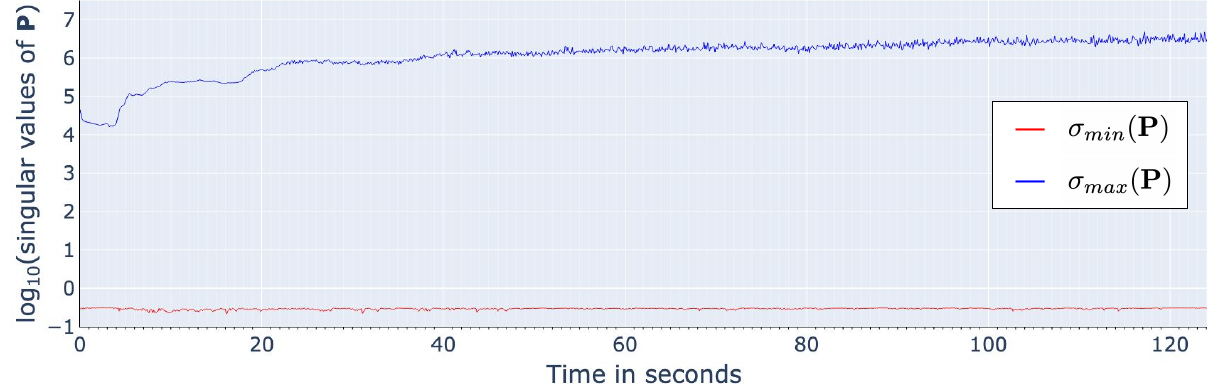}
    \caption{$\sigma_{max}(\mathbf{P})$ and $\sigma_{min}(\mathbf{P})$ over time on a VINS dataset sequence. Note that the states are already re-scaled so that their variances are almost the same, i.e., diagonal elements of $\mathbf{P}$ are approximately equal.}
    \label{fig:singular_value}
    \vspace{-5mm}
\end{figure}

In VINS we have identified two causes for this issue. First, the uncertainties of different states may be of different magnitudes in their chosen units. For instance, it is common for the camera's focal length to have a standard deviation of around one pixel, while the uncertainty of the $^I{\mathbf{p}}_C$ is typically as small as 0.001 meters. Second, the variances of 4 DOF (position and rotation about the gravity vector) in $^G{\mathbf{q}}_{I_k},^G{\mathbf{p}}_{I_k}$ increase over time due to the 4 unobservable directions of VINS (see the observability analysis of~\cite{Hesch2014}) and the accumulation of uncertainty.
%
%Empirically, we have observed that even if the states are re-scaled so that their variances are almost the same, $\sigma_{max}(\mathbf{P})$ keeps increasing while $\sigma_{min}(\mathbf{P})$ remains small (see Fig.~\ref{fig:singular_value}).
%Empirically, we have observed that $\sigma_{max}(\mathbf{P})$ corresponds to a direction highly correlated with the unobservable directions and keeps increasing, while $\sigma_{min}(\mathbf{P})$ corresponds to a highly observable direction and remains small (see Fig.~\ref{fig:singular_value}).
%
As a result, $\sigma_{max}(\mathbf{P})$ increases with the variance of the unobservable directions over time, while $\sigma_{min}(\mathbf{P})$ remains small due to corresponding to a highly observable direction (see Fig.~\ref{fig:singular_value}).
Therefore, $\kappa(\mathbf{P})=\kappa(\mathcal{I})$ also increases with time and the LS solver will become unstable eventually.

Addressing the first issue is relatively straightforward. One can either change the units of the states beforehand or apply a scaling to the states on the fly in order to make the uncertainty of different states approximately equal. The scaling factors can be obtained from the diagonal elements of $\mathcal{I}$.
However, this technique is only effective in reducing $\kappa(\mathcal{I})$ when $\mathcal{I}$ is highly dominated by its diagonal.
Unfortunately, most existing VINS state parameterizations lead to off-diagonal elements of significant magnitudes relative to the diagonal, which is why applying a scaling does not resolve all the issues (Fig.~\ref{fig:singular_value}). Specifically, as in~\eqref{eq:state_pose}, VINS filters typically estimate a sliding window of recent global poses (i.e., w.r.t the global frame $\{G\}$). These poses are all highly correlated with the aforementioned unobservable directions, and thus their cross-covariances are of large magnitudes, resulting in an ill-conditioned $\mathcal{I}$ even after applying a scaling.

Considering that the root cause of this issue is from the global poses, one potential solution is the robocentric state parameterization in~\cite{huai2022robocentric}.
In our notation, it is in the form of
\begin{align*}
    \begin{bmatrix}
    {}^{\scriptscriptstyle I_{k-l+2}}{\mathbf{p}}_{\scriptscriptstyle I_{k-l+1}}^\top\
    {}^{\scriptscriptstyle I_{k-l+2}}_{\scriptscriptstyle I_{k-l+1}}\mathbf{q}^\top\
    \dots\ 
    {}^{\scriptscriptstyle I_k}{\mathbf{p}}_{\scriptscriptstyle I_{k-1}}^\top\
    {}^{\scriptscriptstyle I_k}_{\scriptscriptstyle I_{k-1}}\mathbf{q}^\top\
    {}^{\scriptscriptstyle G}{\mathbf{p}}_{\scriptscriptstyle I_k}^\top\
    {}^{\scriptscriptstyle G}_{\scriptscriptstyle I_k}\mathbf{q}^\top
    \end{bmatrix}^\top
\end{align*}
where only the latest pose is global, while all other past poses are relative poses, i.e., each pose is defined relative to the pose at its next time step.
In contrast to global poses, relative poses are observable and have low correlation with other global poses. Therefore, this parameterization is expected to result in small cross-covariances between the poses in the sliding window, which would enable the scaling technique to effectively reduce $\kappa(\mathcal{I})$. The downside to such parameterization is additional computational overhead;
It increases propagation cost because poses are reparameterized as the window slides, and the visual measurement functions require additional inputs, a chain of relative poses, thereby reducing Jacobian matrix sparsity.

\begin{figure*}
    \centering
    \begin{subfigure}{0.23\textwidth}
        \includegraphics[width=\textwidth]{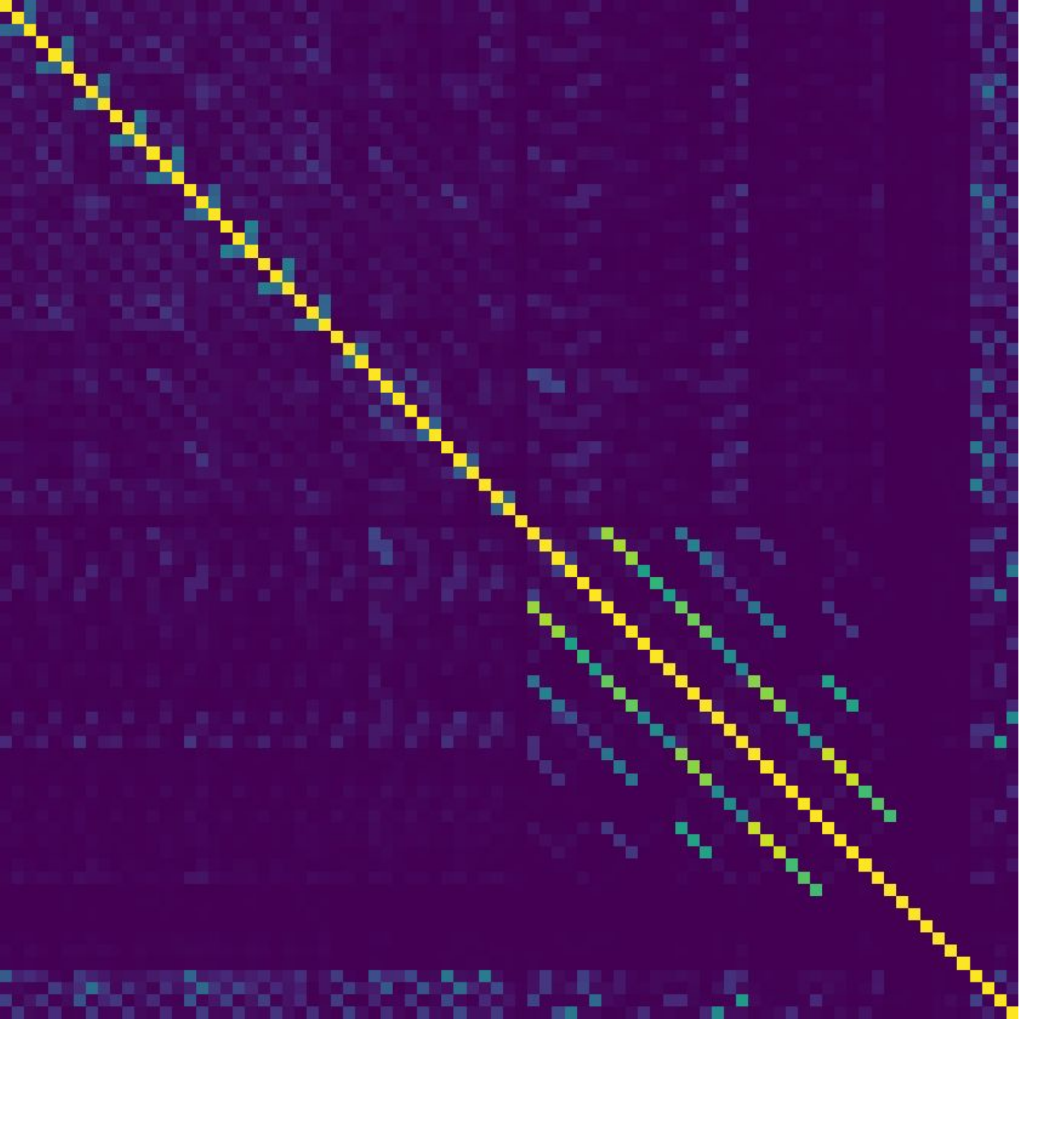}
        \caption{}
        \label{fig:r22_transpose_r22}
    \end{subfigure}
    % \hfill
    \begin{subfigure}{0.23\textwidth}
        \includegraphics[width=\textwidth]{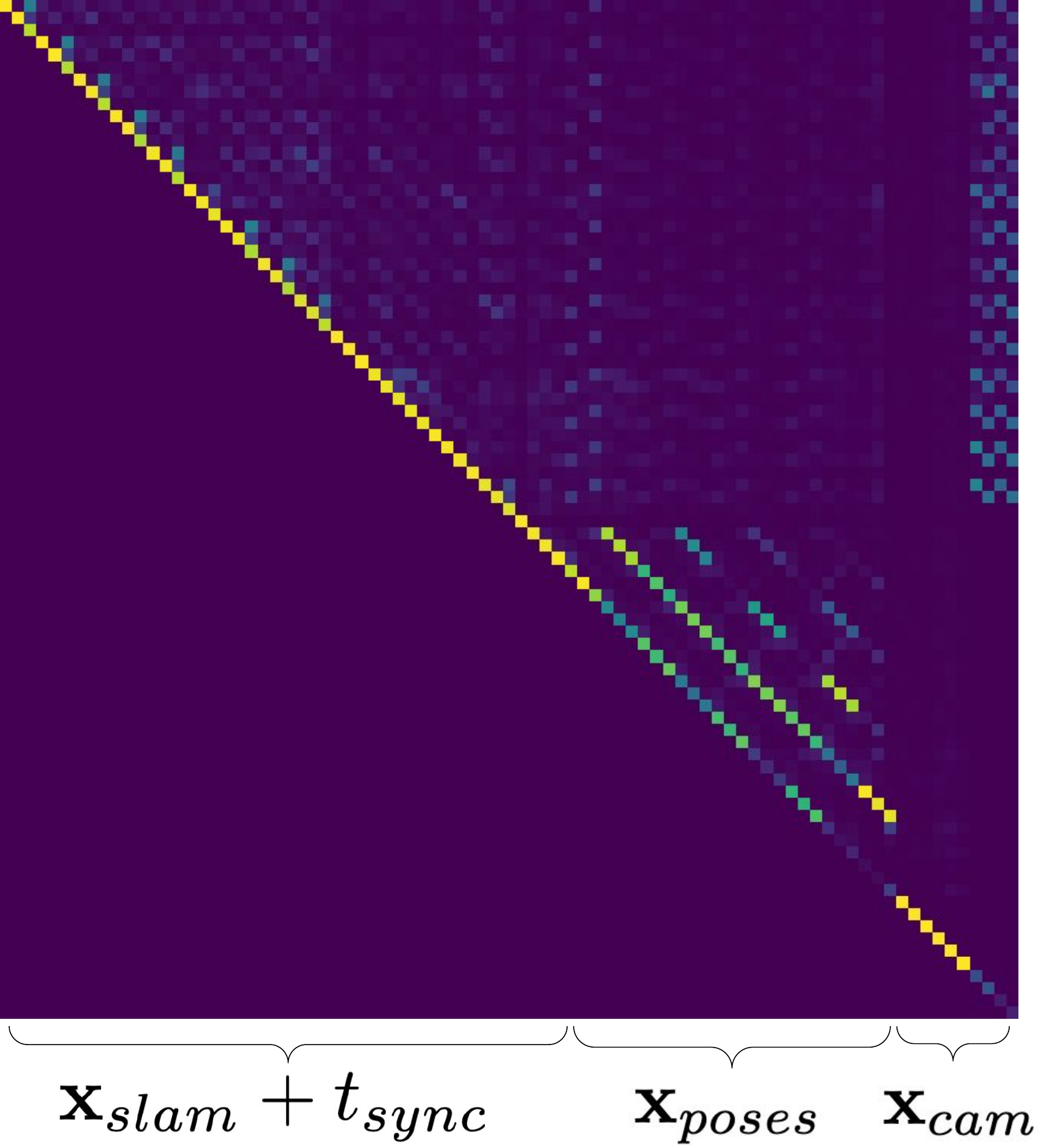}
        \caption{}
        \label{fig:r22}
    \end{subfigure}
    % \hfill
    \begin{subfigure}{0.23\textwidth}
        \includegraphics[width=\textwidth]{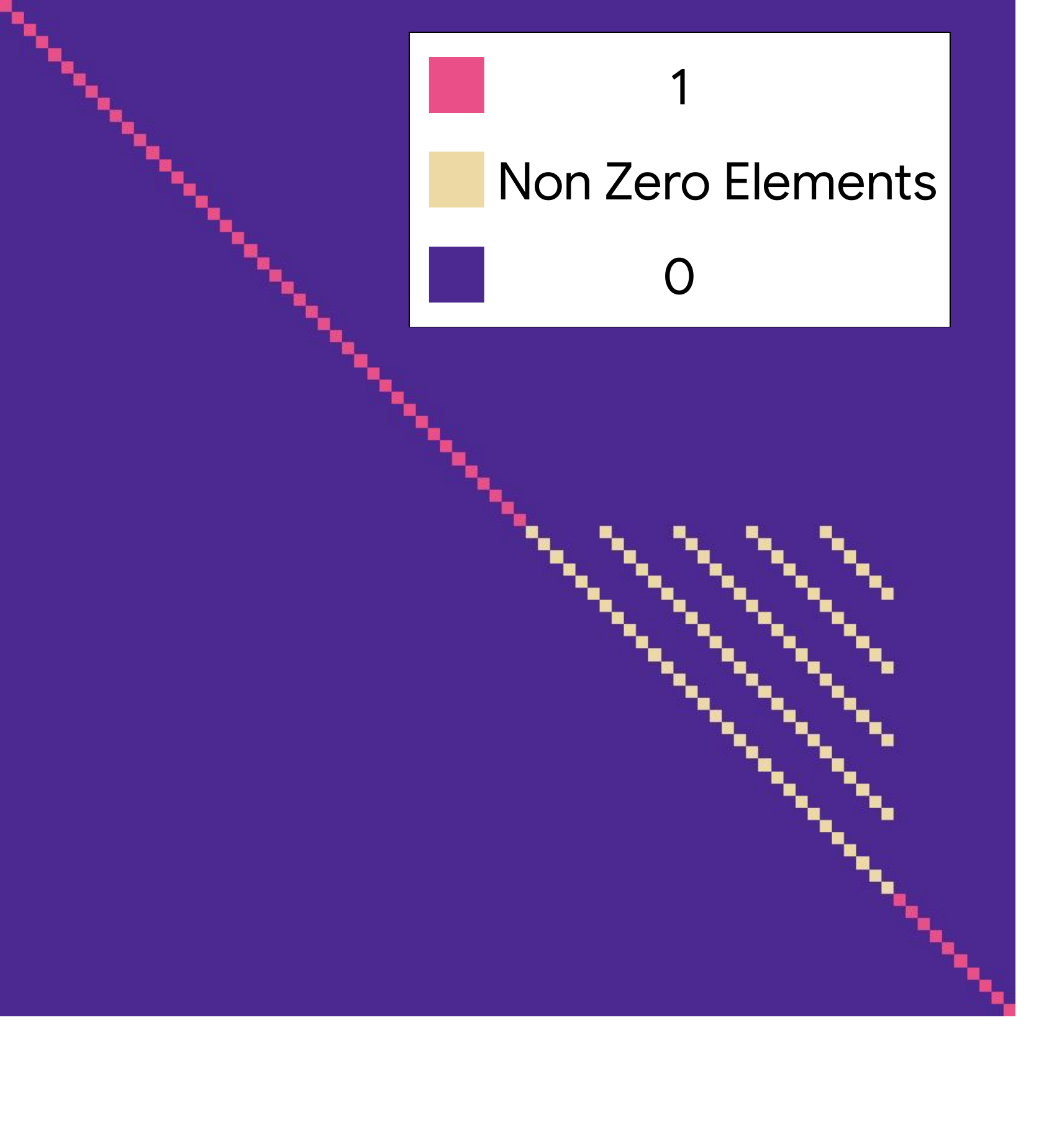}
        \caption{}
        \label{fig:m_spai}
    \end{subfigure}
    \begin{subfigure}{0.23\textwidth}
        \includegraphics[width=\textwidth]{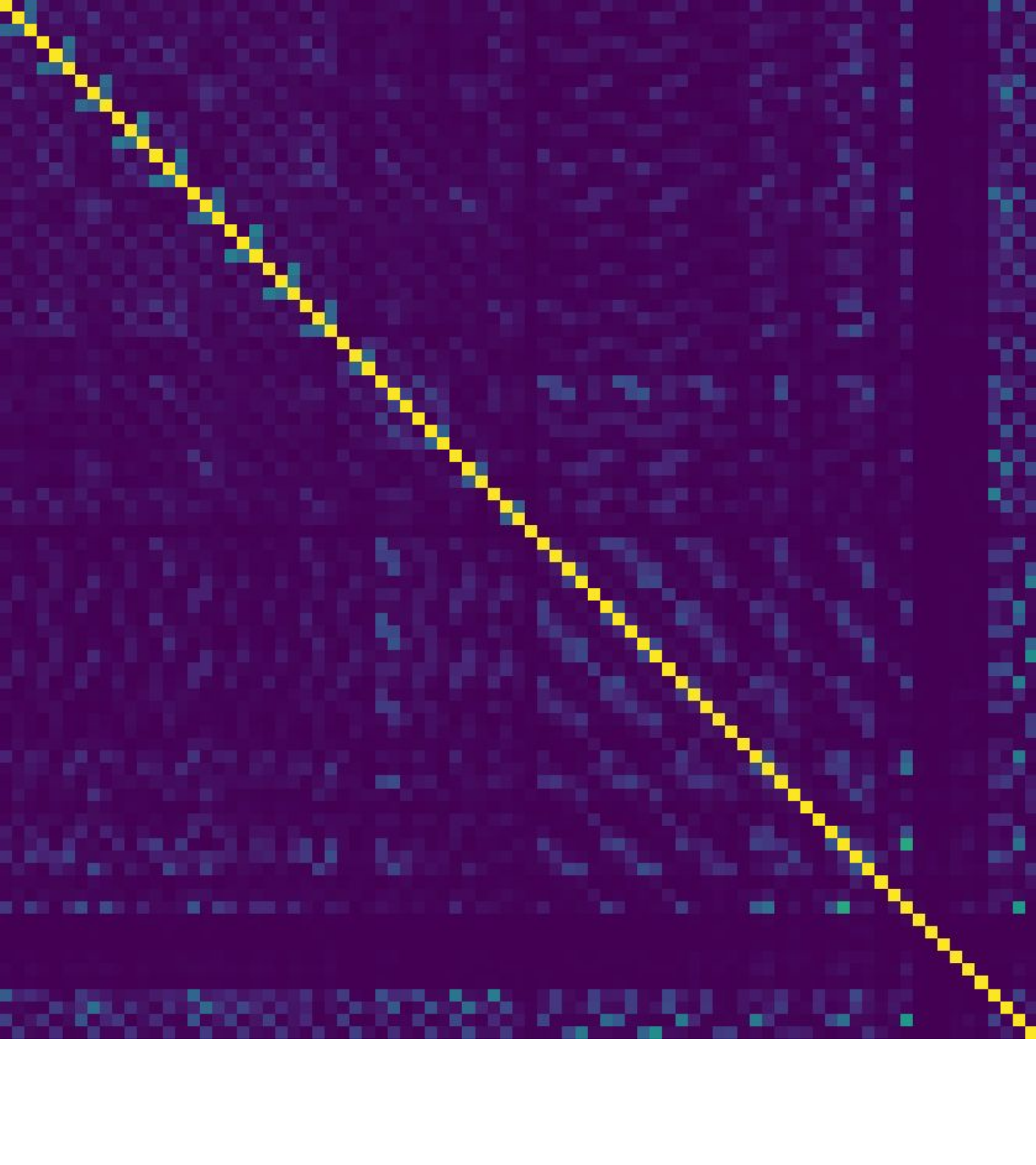}
        \caption{}
        \label{fig:rm_inverse}
    \end{subfigure}
    \begin{subfigure}{0.02\textwidth}
        \includegraphics[scale=0.14]{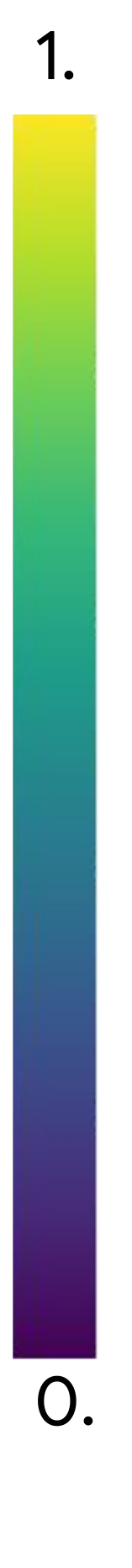}
        \label{fig:scale}
    \end{subfigure}
    \caption{An example of the square root information matrices and the preconditioner. Here, we define a function of matrix, $\mathbf{g}(\mathbf{A})=\mathbf{A}\mathbf{D}^{-1}$, where $\mathbf{D}$ is diagonal and $\mathbf{D}^2=diag(\mathbf{A}^\top\mathbf{A})$. This is essentially applying a scaling, so that the relative magnitude between different elements are not impacted by the original scales of the states.
    a) Magnitude of $\mathbf{g}(\mathbf{R}_{\scriptscriptstyle22})^\top\mathbf{g}(\mathbf{R}_{\scriptscriptstyle22})$. Except for a few correlation elements between poses ($l=5$), almost all off-diagonal elements are approximately zero. b) Magnitude of $\mathbf{g}(\mathbf{R}_{\scriptscriptstyle22})$, exhibiting a similar pattern. We construct a sparse approximation of it as $\mathbf{M}_{\scriptscriptstyle SPAI}$. c) Sparsity pattern of $\mathbf{M}_{\scriptscriptstyle SPAI}$. d) Magnitude of $\mathbf{g}(\mathbf{R}_{\scriptscriptstyle22s})^\top\mathbf{g}(\mathbf{R}_{\scriptscriptstyle22s})$, which is almost an identity matrix, suggesting our preconditioner is effective.}
    \label{fig:factor_spai}
\end{figure*}

To effectively reduce the condition number while minimizing the overhead, we propose to retain the global pose sliding window in~\eqref{eq:state_pose} as it is but employ the preconditioners in PC-SRIF (see Sec.~\ref{sec:estimator}) to address the conditioning issues, which can be viewed as applying a temporary linear reparameterization while solving the LS. Since we only apply them during the update step, no overhead is introduced to other components.
Moreover, when using (PC-)SRIF, $\mathbf{R}_{\scriptscriptstyle22}$ in~\eqref{eq:srif_small1} is almost a sparse matrix (see Fig.~\labelcref{fig:r22_transpose_r22,fig:r22}), which allows for the efficient construction and application of a sparse approximation of inverse (SPAI) preconditioner.
In summary, we leverage Jacobi and SPAI preconditioners to address the aforementioned issues and enable the Cholesky LS solver in float32 for VINS, which are detailed below.

First, note that PC-SRIF only needs to focus on the condition number of $\mathbf{R}_{\scriptscriptstyle22}^\oplus$, a smaller matrix than the $\mathcal{I}^\oplus$ or $\mathbf{R}^\oplus$ that needs to be considered in IF. This frees us from dealing with other states correlated with the unobservable directions but not involved in the visual measurement model such as $^{\scriptscriptstyle G}\mathbf{v}_{\scriptscriptstyle I_k}$.
Second, although the goal is to find a matrix $\mathbf{M}$ such that $\kappa(\mathbf{R}_{\scriptscriptstyle22}^\oplus\mathbf{M}^{-1})$ is small, to limit the cost, we use information already available before constructing $\mathbf{R}_{\scriptscriptstyle22}^\oplus$. Thus, we find preconditioners for $\mathbf{R}_{\scriptscriptstyle22}$ as an approximation. As demonstrated by the experiments, this approach is adequate for substantially reducing the condition number in VINS.
%As described above, the high condition number of the hessian can be attributed to two primary factors: the prevalence of dominant off-diagonal terms associated with cross-correlations of poses and scale of different quantities in the state vector.  Leveraging this knowledge, we can approximate the Hessian structure sparsely to create preconditioners. In PC-SRIF we deploy the following SPAI preconditioners, $M_{decorrelate}$ and $M_{\scriptscriptstyle Jacobi}$ in the given order: 

Specifically, we first apply a SPAI preconditioner $\mathbf{M}_{\scriptscriptstyle SPAI}$ (see Fig.~\ref{fig:m_spai}), defined as
\begin{align*}
&P=\{ \tilde{\mathbf{x}}_{pose} \mid \tilde{\mathbf{x}}_{pose} \text{ is the error state of a pose in }\mathbf{x}_{poses}\}\\
&S=\{ (pos(\tilde{\mathbf{x}}_{i}) + k, pos( \tilde{\mathbf{x}}_{j}) + k) \mid \tilde{\mathbf{x}}_{i}, \tilde{\mathbf{x}}_{j} \in P,k \in [0, 5] \}\\
&\mathbf{M}_{\scriptscriptstyle SPAI}(i,j) =
    \begin{cases}
        \mathbf{R}_{\scriptscriptstyle22}(i,j) & \text{if } (i,j) \in S \\
        1 & \text{if } i = j,(i,j) \notin S \\
        0 & \text{else}
    \end{cases}
\end{align*}
where we use $\mathbf{\tilde{x}}$ to denote the error state of a state $\mathbf{x}$ as~\cite{Wu2015} and $pos(\tilde{\mathbf{x}})$ means the index of an error state $\tilde{\mathbf{x}}$ in $\tilde{\mathbf{x}}_{state}$.
Intuitively, we have observed that the dominant off-diagonal elements are from the correlation between the poses, which aligns with our conditioning analysis, and thus $\mathbf{M}_{\scriptscriptstyle SPAI}$ keeps those elements between poses while ignoring other off-diagonal elements.
Then, a Jacobi preconditioner is applied to the preconditioned $\mathbf{R}_{\scriptscriptstyle22}$, $\mathbf{R}_{\scriptscriptstyle22s}=\mathbf{R}_{\scriptscriptstyle22}\mathbf{M}_{\scriptscriptstyle SPAI}^{-1}$, which alleviates the conditioning issues introduced by the different scales of the states. Formally, $\mathbf{M}_{\scriptscriptstyle Jacobi}$ is given by
\begin{equation}
\mathbf{M}_{\scriptscriptstyle Jacobi}(i,j) =
    % \begin{cases}
    %     \lVert\mathbf{R}_{\scriptscriptstyle22s}(:, i) \rVert _{2} & \text{if } i = j \\
    %     0 & \text{else}
    % \end{cases}
    \lVert\mathbf{R}_{\scriptscriptstyle22s}(:, i) \rVert _{2} \text{ if } i = j \text{ else } 0
\end{equation}
and the corresponding $\mathbf{M}$ in \eqref{eq:normal_eq_small_pc} equals $\mathbf{M}_{\scriptscriptstyle Jacobi}\mathbf{M}_{\scriptscriptstyle SPAI}$.

Note that $\mathbf{M}_{\scriptscriptstyle Jacobi}$ and $\mathbf{M}_{\scriptscriptstyle SPAI}$ are matrices with known sparse patterns, and thus can be applied without explicit construction or computing inversion.
Consequently, our preconditioning introduces minor computational cost to the LS solver, which we demonstrate in the next section.
%It is crucial to emphasize that the preconditioners only consider the prior square root information matrix $\mathbf{R}$, but not the Jacobian matrix $\mathbf{H}$. Consequently, they do not directly enhance the conditioning of the posterior information matrix $\mathbf{R}^\oplus$, the matrix inverted when solving the LS problem. Nevertheless, as demonstrated by the experimental findings, this approach is adequate for substantially reducing the condition number in VINS. Additionally, both preconditioners can be constructed and applied without the explicit formation of the preconditioner matrices or the computation of the inverse. This enables the implementation of both preconditioners without introducing significant computational complexity, resulting in only a marginal overhead compared to the LS solver itself.
%
%It is important to note that the preconditioners employed do not incorporate any of the Jacobian terms, resulting in a sparse approximation of the Hessian. This is mainly to avoid large processing cost considering the fact that the dimensions of J can be exceptionally large. Additionally, both the preconditioners can be constructed and applied without explicitly constructing the hessian or computing the inverse. Thus both the preconditioners can be implemented without incurring significant computational complexity. Furthermore, the overhead of $M$ is marginal compared to the benefits from using Cholesky solver instead of QR thereby making PC-SRIF more efficient compared to SRIF. 
\section{Experimental Results}
\subsection{Experiment Setup}
The goal of our experiments is to provide a comparative analysis of various filtering-based VINS estimators and validate the theoretical advantages of PC-SRIF on practical platforms with limited computational resources. To this end, we implemented a VINS with three estimator options: KF, SRIF, and PC-SRIF. Except for the estimator, all other modules such as image processing and measurement function evaluation are shared, ensuring a fair comparison among the estimators themselves.
In this VINS, the number of poses in the sliding window is set to $l=11$, and at most 15 SLAM features and 35 MSCKF features (as defined in~\cite{Wu2015}) are processed at each time step.
Under this setup, the maximum measurement dimension $m$ is $995$, and the state size in the update $n_{\scriptscriptstyle2}$ is 122.
Note that since all these estimators are mathematically equivalent, while they may exhibit different efficiency, the only source of difference in their estimates is numerical error.

We evaluate our VINS on a cell phone dataset collected on a Pixel 3.
Additionally, we include the results of a state-of-the-art open source VINS, OpenVINS~\cite{geneva2020openvins} (a float64 KF), on monocular EuRoC dataset~\cite{Burri2016}.
%VINS-Mono~\cite{Qin2018}. VINS-Mono's estimator can be viewed as a float64 IF with iterated updates similar to the iterated extended Kalman filter, and thus it is expected to be slower but with satisfactory accuracy.
All systems are executed on a Pixel 7 phone and affiliated to one 2.85 GHz Cortex-X1 CPU. The absolute trajectory error (ATE) and relative trajectory error (RTE, with 1 sec interval) are employed for accuracy comparison.
%Note that given our focus on real-time applications, the pose error for each timestamp is computed using the first real-time pose estimate reported by the estimators, rather than any refined ones.

%In the subsequent subsections, we first present a numerical stability analysis, which demonstrates the effectiveness of our preconditioners in reducing the condition number and PC-SRIF's numerical stability. Then, we evaluate the end-to-end accuracy of the PC-SRIF, and the estimator runtime is measured to showcase its superior efficiency.
The following subsections will first present a numerical stability analysis, demonstrating how our preconditioners ensure PC-SRIF's numerical stability by reducing the condition number. Subsequently, we will evaluate the end-to-end accuracy of the PC-SRIF to confirm that its efficiency gains do not compromise accuracy. Finally, the estimator runtime will be measured to showcase its superior efficiency.

\subsection{Numerical Stability Analysis}
\begin{figure}
    \centering
    \includegraphics[width=0.48\textwidth]{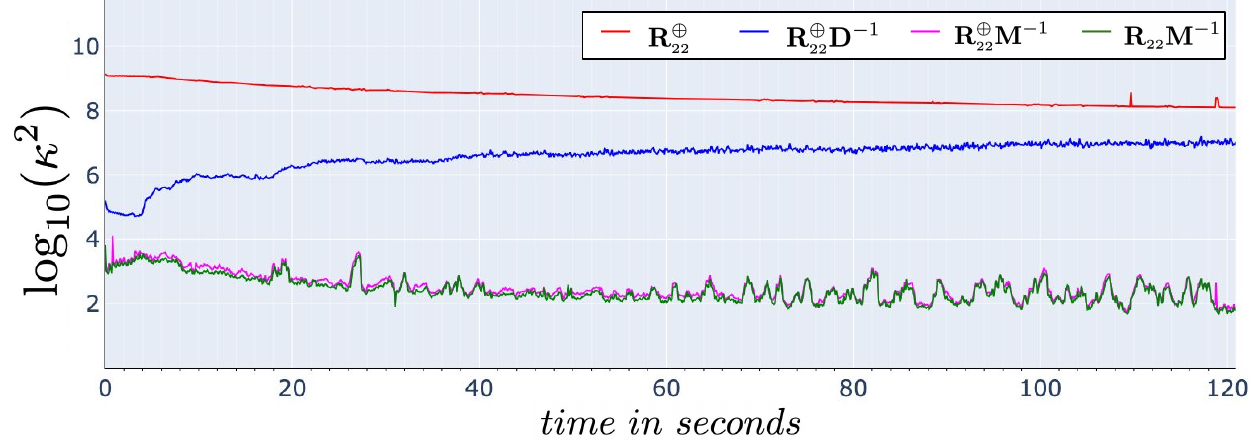}
    \caption{Condition number of square root information matrices over time from running our PC-SRIF on a VINS dataset sequence. $\mathbf{D}$ is a diagonal matrix such that $\mathbf{D}^2=diag(\mathbf{R}_{\scriptscriptstyle22}^{\oplus\top}\mathbf{R}_{\scriptscriptstyle22}^\oplus)$. $\kappa^2(\mathbf{R}_{\scriptscriptstyle22}^\oplus)>>1/\epsilon_{float32}\approx8.4\times10^6$, and thus the Cholesky solver is unstable for the original LS. On contrary, we can use it for the preconditioned system, since $\kappa^2(\mathbf{R}_{\scriptscriptstyle22}^\oplus\mathbf{M}^{-1})<<1/\epsilon_{float32}$.}
    \label{fig:cond_r22}
    \vspace{-1mm}
\end{figure}

The key factor if a Cholesky LS solver can be applied stably is the condition number of the (square root) information matrix given a machine epsilon~\cite{golub2012matrix}.
As discussed earlier, in solving in LS \eqref{eq:srif_small1} of SRIF, only a sub-block $\mathbf{R}^\oplus_{\scriptscriptstyle22}$ of the square root information matrix needs to be inverted.
Nevertheless, as shown in Fig.~\ref{fig:cond_r22}, with the global pose parameterization, $\kappa^2(\mathbf{R}_{\scriptscriptstyle22}^\oplus)$ can be as high as $10^9$, which necessitates float64 for the Cholesky LS solver.
$\kappa^2(\mathbf{R}_{\scriptscriptstyle22}^\oplus\mathbf{D}^{-1})$ is still high in float32 and increases with time, suggesting that only applying a diagonal scaling does not resolve the issue.
In contrast, in PC-SRIF, the preconditioned square root information matrix $\mathbf{R}_{\scriptscriptstyle22}\mathbf{M}^{-1}$ and $\mathbf{R}^\oplus_{\scriptscriptstyle22}\mathbf{M}^{-1}$ are constantly well-conditioned,
This evidences two conclusions: (i) Our proposed VINS preconditioners are effective for enabling the Cholesky solver in float32. (ii) For constructing VINS preconditioners, it is sufficient to consider only the prior information matrix $\mathbf{R}_{\scriptscriptstyle22}$ rather than the posterior $\mathbf{R}_{\scriptscriptstyle22}^\oplus$.

\subsection{Accuracy and Efficiency}
As mentioned before, PC-SRIF differs from KF and SRIF only in numerical difference, and thus we first compare VINS' accuracy to show PC-SRIF's numerical stability in float32.
As shown in Table~\ref{tab:ate},\ref{tab:ate2}, all the estimators implemented by us, regardless in float32 and float64, achieve similar accuracy, demonstrating that they are all sufficiently stable in float32.
PC-SRIF does not compromise accuracy compared to SRIF even if it employs the less stable Cholesky solver, attributed to its improved conditioning.
In addition, they are comparable with OpenVINS, indicating that our VINS achieves reasonable accuracy for comparing efficiency.

Next, we evaluate the computational cost of the considered estimators to demonstrate PC-SRIF's superior efficiency.
%
%Table~\ref{tab:time} summarizes the time spent on the estimators, i.e., linear algebra operations in propagation, marginalization, and update, excluding other modules such as image processing (4.65 ms), measurement function/Jacobian evaluation and bookkeeping (1.96 ms).
Table~\ref{tab:time} summarizes the time spent on the estimators, i.e., linear algebra operations in propagation, marginalization, and update, excluding other modules such as image processing, measurement function evaluation, and bookkeeping.
Though KF's marginalization is trivial, thanks to our novel algorithm, (PC-)SRIF's marginalization cost is minor compared to other steps.
As expected, the update step is the most expensive for all, while PC-SRIF makes it faster by 62\% compared to SRIF.
In addition, out of the 0.53 ms in the update, the time spent on the preconditioning is only 0.08 ms, a minor overhead.
Overall, KF is much slower than SRIF due to its expensive update, agreeing with \cite{Wu2015}'s results.
Remarkably, PC-SRIF is even faster than SRIF by 41\%, which validates the theoretical advantages in our previous analysis.

\begin{table}
    \centering
    \captionsetup{font=footnotesize}
    \caption{EuRoC dataset (11 sequences): Mean ATE (top 2 rows) and RTE (bottom 2 rows) in meters/degrees}
    {\scriptsize
    \begin{tabular}{cccccc}
    \hline
     & KF & SRIF & PC-SRIF & OpenVINS \\ \hline
    float32 & 0.16 / 1.70 & 0.16 / 1.65 & 0.16 / 1.64 & - \\
    float64 & 0.16 / 1.69 & 0.17 / 1.65 & 0.16 / 1.66 & 0.14 / 1.53 \\\hline
    float32 & 0.046 / 0.33 & 0.045 / 0.31 & 0.045 / 0.31 & - \\
    float64 & 0.047 / 0.33 & 0.046 / 0.31 & 0.045 / 0.31 & 0.042 / 0.30 \\\hline
    \end{tabular}
    }
    \label{tab:ate}
\end{table}
\begin{table}
    \centering
    \captionsetup{font=footnotesize}
    \caption{Cell phone dataset: Mean ATE (top 2 rows) and RTE (bottom 2 rows) in meters/degrees}
    {\scriptsize
    \begin{tabular}{cccccc}
    \hline
     & KF & SRIF & PC-SRIF \\ \hline
    float32 & 0.11 / 2.16 & 0.11 / 1.98 & 0.10 / 2.00 \\
    float64 & 0.10 / 2.24 & 0.10 / 1.82 & 0.11 / 2.02 \\\hline
    float32 & 0.024 / 0.31 & 0.022 / 0.27 & 0.021 / 0.27 \\
    float64 & 0.024 / 0.30 & 0.021 / 0.29 & 0.022 / 0.27 \\\hline
    \end{tabular}
    }
    \label{tab:ate2}
\end{table}
\begin{table}
    \centering
    \captionsetup{font=footnotesize}
    \caption{Cell phone dataset: Average run time (ms)}
    {\scriptsize
    \begin{tabular}{cccccc}
    \hline
    & KF float32 & SRIF float32 & PC-SRIF float32 \\ \hline
    Propagation & 0.29 & 0.19 & 0.19 \\
    Marginalization & 0.00 & 0.07 & 0.07  \\
    Update & 1.56 & 0.86 & 0.53  \\\hline
    Estimator Total & 1.85 & 1.12 & \textbf{0.79} \\
    \end{tabular}
    }
    \label{tab:time}
\end{table}
\FloatBarrier
\section{Conclusions and Future Work}
In conclusion, this paper introduces PC-SRIF, a novel VINS estimator characterized by high efficiency and numerical stability in float32. PC-SRIF combines the efficiency advantages of IF and SRIF, most notably a fast and stable LS solver. Existing works typically employ the more stable QR LS solver in float32 or the faster Cholesky LS solver in float64, while the Cholesky LS solver in float32 is deemed unstable due to the ill-conditioned information matrix in VINS. In contrast, PC-SRIF enables the utilization of the Cholesky LS solver in float32, outperforming alternative options. This achievement is attributed to analyzing the root cause of the conditioning issues in VINS and leveraging an effective preconditioning technique. Experimental results substantiate the effectiveness of the proposed preconditioners and the efficiency advantages of PC-SRIF over alternative estimators, KF and SRIF.

It is noteworthy that the design of the proposed preconditioners stems from the observation that the square root information matrix in our VINS is dominated by a small subset of elements (Fig.~\ref{fig:factor_spai}). This implies that computations involving this matrix can potentially be substituted with their sparse approximations without compromising accuracy significantly. As future work, we aim to leverage this sparsity in VINS to further enhance the estimator's efficiency.

%\clearpage
%\columnbreak
%\afterpage{\clearpage}
%\addtolength{\textheight}{-12cm}   % This command serves to balance the column lengths
                                  % on the last page of the document manually. It shortens
                                  % the textheight of the last page by a suitable amount.
                                  % This command does not take effect until the next page
                                  % so it should come on the page before the last. Make
                                  % sure that you do not shorten the textheight too much.

%%%%%%%%%%%%%%%%%%%%%%%%%%%%%%%%%%%%%%%%%%%%%%%%%%%%%%%%%%%%%%%%%%%%%%%%%%%%%%%%

%%%%%%%%%%%%%%%%%%%%%%%%%%%%%%%%%%%%%%%%%%%%%%%%%%%%%%%%%%%%%%%%%%%%%%%%%%%%%%%%

\bibliographystyle{IEEEtran} % style of bibliography
\bibliography{ref, IEEEabrv}
%%%%%%%%%%%%%%%%%%%%%%%%%%%%%%%%%%%%%%%%%%%%%%%%%%%%%%%%%%%%%%%%%%%%%%%%%%%%%%%%
% \section*{APPENDIX}

% Appendixes should appear before the acknowledgment.

% \section*{ACKNOWLEDGMENT}

% The preferred spelling of the word ÒacknowledgmentÓ in America is without an ÒeÓ after the ÒgÓ. Avoid the stilted expression, ÒOne of us (R. B. G.) thanks . . .Ó  Instead, try ÒR. B. G. thanksÓ. Put sponsor acknowledgments in the unnumbered footnote on the first page.

% %%%%%%%%%%%%%%%%%%%%%%%%%%%%%%%%%%%%%%%%%%%%%%%%%%%%%%%%%%%%%%%%%%%%%%%%%%%%%%%%

% References are important to the reader; therefore, each citation must be complete and correct. If at all possible, references should be commonly available publications.

\end{document}